\documentclass[10pt, conference]{IEEEtran}
\IEEEoverridecommandlockouts
\usepackage{cite}
\usepackage{subcaption}
\usepackage{amsmath,amssymb,amsfonts} 
\usepackage{mathtools}
\usepackage{algorithmic}
\usepackage{graphicx}
\usepackage{textcomp}
\usepackage[dvipsnames]{xcolor}
\usepackage{comment}
\usepackage{multirow}
\usepackage{array}
\usepackage{tabularx}
\usepackage[colorlinks]{hyperref}
\def\BibTeX{{\rm B\kern-.05em{\sc i\kern-.025em b}\kern-.08em
    T\kern-.1667em\lower.7ex\hbox{E}\kern-.125emX}}

\usepackage{listings}
\usepackage{xcolor}

\colorlet{punct}{red!60!black}
\definecolor{background}{HTML}{EEEEEE}
\definecolor{delim}{RGB}{20,105,176}
\colorlet{numb}{magenta!60!black}

\lstdefinelanguage{json}{
    basicstyle=\tiny\ttfamily,
    showstringspaces=false,
    breaklines=true,
    backgroundcolor=\color{background},
    literate=
     *{0}{{{\color{numb}0}}}{1}
      {1}{{{\color{numb}1}}}{1}
      {2}{{{\color{numb}2}}}{1}
      {3}{{{\color{numb}3}}}{1}
      {4}{{{\color{numb}4}}}{1}
      {5}{{{\color{numb}5}}}{1}
      {6}{{{\color{numb}6}}}{1}
      {7}{{{\color{numb}7}}}{1}
      {8}{{{\color{numb}8}}}{1}
      {9}{{{\color{numb}9}}}{1}
      {:}{{{\color{punct}{:}}}}{1}
      {,}{{{\color{punct}{,}}}}{1}
      {\{}{{{\color{delim}{\{}}}}{1}
      {\}}{{{\color{delim}{\}}}}}{1}
      {[}{{{\color{delim}{[}}}}{1}
      {]}{{{\color{delim}{]}}}}{1},
}

\graphicspath{{media/}}

\DeclareMathOperator{\part}{part}

\newsavebox{\twosubbox}

\newcommand\myshade{65}
\colorlet{mylinkcolor}{violet}
\colorlet{mycitecolor}{YellowOrange}
\colorlet{myurlcolor}{Aquamarine}

\hypersetup{
  linkcolor  = mylinkcolor!\myshade!black,
  citecolor  = mycitecolor!\myshade!black,
  urlcolor   = myurlcolor!\myshade!black,
  colorlinks = true,
}
    
\begin{document}

\title{CLS-CAD: Synthesizing CAD Assemblies in Fusion 360\\

\thanks{Funded by the Deutsche Forschungsgemeinschaft (DFG, German Research Foundation) – Project Number 276879186}
}

\author{\IEEEauthorblockN{Constantin Chaumet}
\IEEEauthorblockA{\textit{Department of Computer Science} \\
\textit{TU Dortmund University}\\
Dortmund, North Rhine-Westphalia, Germany \\
ORCID: 0000-0002-8359-8030}
\and
\IEEEauthorblockN{Jakob Rehof}
\IEEEauthorblockA{\textit{Department of Computer Science} \\
\textit{TU Dortmund University}\\
Dortmund, North Rhine-Westphalia, Germany \\
jakob.rehof@tu-dortmund.de}
}

\maketitle
\begin{abstract}
The CAD design process includes a number of repetitive steps when creating assemblies. This issue is compounded when engineering whole product lines or design families, as steps like inserting parts common to all variations, such as fasteners and product-integral base parts, get repeated numerous times. This makes creating designs time-, and as a result, cost-intensive. While many CAD software packages have APIs, the effort of creating use-case specific plugins to automate creation of assemblies usually outweighs the benefit.

We developed a plugin for the CAD software package ``Fusion 360'' which tackles this issue. The plugin adds several graphical interfaces to Fusion 360 that allow parts to be annotated with types, subtype hierarchies to be managed, and requests to synthesize assembly programs for assemblies to be posed. The plugin is use-case agnostic and is able to generate arbitrary open kinematic chain structures. We envision engineers working with CAD software being able to make designed parts reusable and automate the generation of different design alternatives as well as whole product lines. 
\end{abstract}

\begin{IEEEkeywords}
Combinatory Logic, Synthesis, Engineering, Robotics, CAD, Autodesk Fusion 360, Knowledge-Based Engineering
\end{IEEEkeywords}

\section{Introduction}
When designing a physical product, an integral part of the process is modeling the product in CAD software, so that the necessary parts can be produced later on. 
Usually, a product will consist of many different parts, some off-the-shelf (e.g.\ screws), and some specifically designed for the product. 
It is common practice to design each part in a separate document, so that the individual history and design revisions of the parts get preserved. 
The actual product is then located in a so-called assembly, where the individual parts are inserted and connected to each other. 
Assemblies structure the parts, define movement between them, and allow the product to be thoroughly evaluated before it is produced (including load simulation, thermal simulation, inverse kinematics).

With the ongoing shift of production towards highly customized products in wake of Industry 4.0~\cite{AHELEROFF2021101438}, being able to provide a higher degree of customization is a competitive advantage~\cite{AHELEROFF20191394}. However, due to competition, this must also be possible without significantly increasing costs~\cite{AHELEROFF2021101438}.
An approach to achieve this is knowledge-based engineering. 
Knowledge-based engineering aims to capture and reuse knowledge about the design process and aims to reduce development time and costs as a result~\cite{kbe2}. 
Incorporating knowledge-based engineering specifically into the CAD software stage of the design process can reduce costs, development time and exhibit a positive ROI~\cite{Corallo2009}.

Fusion 360 and the older ``Autodesk Inventor'' share many features necessary to apply knowledge-based engineering~\cite{gembarski}. 
Our developed plugin enhances Fusion 360 with knowledge-based engineering capabilities by making use of the features present in the API.
Creating a good CAD assembly, or CAD model in general, is a trial-and-error process, and recipes for perfect solutions arguably do not exist~\cite{AMADORI2012180}. 
Our plugin features many interactive graphical interfaces tightly integrated into Fusion 360. 
This makes it easy to iteratively improve the synthesized results. 
The plugin is fully integrated with the cloud-based nature of Fusion 360, and allows collaboratively applying knowledge-based engineering to the CAD design of products. 
The plugin allows:\\

\begin{tabularx}{0.45\textwidth}{l X}
   $\bullet$ \textbf{Subtyping.} & Management of three subtype hierarchies to annotate elements with: Formats, Parts, and Attributes.\\
   $\bullet$ \textbf{Annotating.} & Annotation of types to elements in Fusion 360.\\
   $\bullet$ \textbf{Requesting.} & Specification of requests for assemblies.\\ 
   $\bullet$ \textbf{Assembling.} & Automated creation of synthesized assemblies in Fusion 360.\\
\end{tabularx} 
\\

We will briefly cover each of these aspects in the following. 
After that, we will present an industrially significant example, by using our plugin to create fully-functional assemblies of robotic arms.

\begin{figure*}
\centering
\scalebox{1}[1]{
\subfloat[Display of subtype hierarchy for editing.]{\label{fig:typinga}\includegraphics[height=4.7cm]{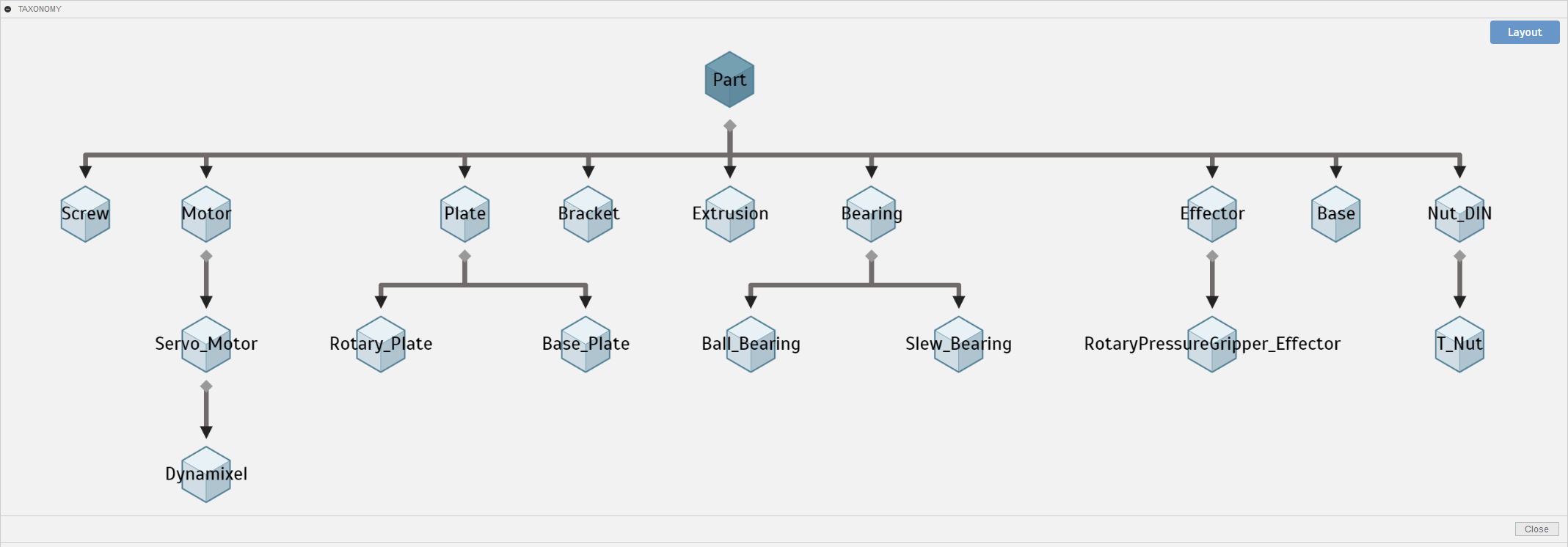}}
}
\hfil
\scalebox{1}[1]{
\subfloat[Display of subtype hierarchy for selecting.]{\label{fig:typingb}\includegraphics[height=4.7cm]{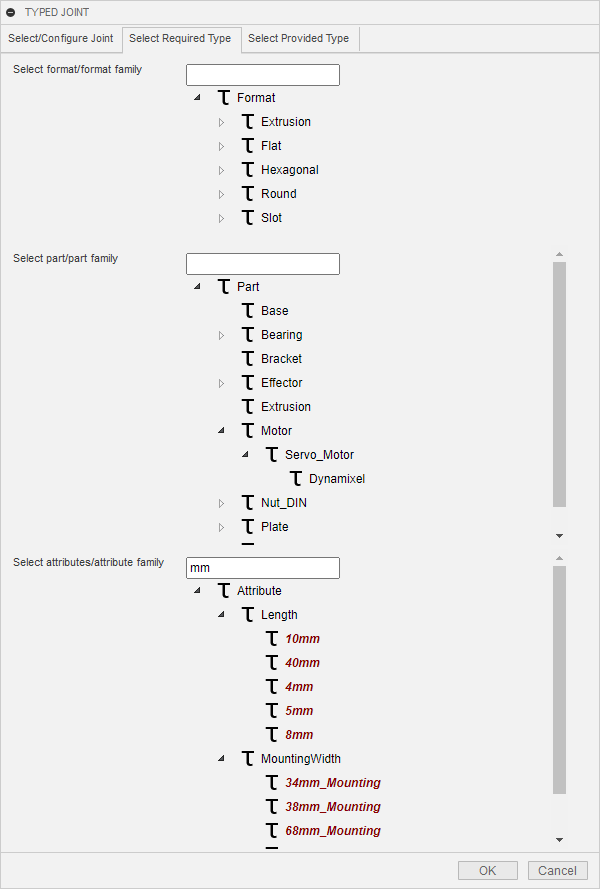}}
}
\caption{Windows to manage the subtype hierarchy, created by the plugin natively in Fusion 360.}
\label{fig:types}
\end{figure*}

\section{Subtyping}
The three subtype hierarchies manage different aspects of typing a part. 
Types in the ``formats'' hierarchy characterize the physical geometry of a connection to be made and thus if correctly applied ensure the correctness of the resulting assemblies in terms of mating geometry, e.g.\ a 3mm hole will not connect to anything that is too large or too small to fasten. 
Types in the ``parts'' hierarchy characterize the intent of a part. 
This allows discerning connections beyond just their geometry, as usually an assembly will need a specific kind of part to be connected to be considered sensible. For instance, a screw could connect to a motor shaft, however, attaching an output bracket or plate makes far more sense.
Types in the ``attributes'' hierarchy characterize any remaining important or interesting information the user might want to specify or query for. 

Managing subtype hierarchies is an integral aspect of the underlying approach (type-oriented synthesis based on combinatory logic~\cite{Bessai:19b}) to synthesize results. 
The approach is implemented by the CLS framework\footnote{https://github.com/tudo-seal/bcls-python}.
The subtype hierarchy directly affects these results, e.g.\ if requesting a screw, the subtype hierarchy decides whether a steel screw is also considered to be a screw. 

The developed plugin adds additional commands to Fusion 360 which open custom graphical user interfaces. 
Two of these graphical interfaces allow managing the three distinct subtype hierarchies, which are utilized to annotate the CAD parts, preparing them for synthesis.
The two interfaces' design is different, as seen in \autoref{fig:types}, to accommodate different use-cases. 

The interface shown in \autoref{fig:typinga} is intended to be used to create and refactor hierarchies.
It allows all CRUD operations to be performed on the hierarchies. 
It can be opened from three separate commands, with each of them displaying a different subtype hierarchy.

The interface shown in \autoref{fig:typingb} is intended to select and assign types to elements in the CAD software. 
It also supports all CRUD operations, however, it restricts these by enforcing that no operation leads to a duplicate name. 
This is done to avoid the user unintentionally merging parts of the tree that were not intended to be merged.

\section{Annotating}
While the presence of types ensures that \textit{combinators} for synthesis can be derived, this is not sufficient to create a physical assembly.
Combinators are objects that have a type and contain code that defines how to construct artifacts if they are applied to each other and interpreted.
For instance, interpreting a combinator \texttt{A} may yield a fragment of code, and the application of a combinator \texttt{B} to \texttt{A} will yield the code of \texttt{B} with the code from \texttt{A} suitably inserted (with arbitrary sophistication and not necessarily monolithically).
A combinator's type directs synthesis by specifying which types are required for using the combinator, and which type it will provide in return. 

To create a physical assembly, types need to be annotated to the actual geometry of parts in a fashion that allows deriving a homogeneous transformation~\cite[Sec.~1.2.3]{HandbookOfRobo}. 
This necessity to derive transformations is well known and analogous to the Denavit-Hartenberg notation in robotics, which is used to define movement and position of individual links in a kinematic chain to each other~\cite{denavit1955kinematic}. 
Links are assumed as rigid, structural components (or in the context of an assembly, a group of parts forming such a component)~\cite[Sec.~1.3]{HandbookOfRobo}.

To suitably annotate types to the actual geometry, our plugin applies a concept from Fusion 360's API, ``Attributes''\footnote{https://help.autodesk.com/view/fusion360/ENU/?guid=GUID-BAF017FE-10B8-4612-BDE2-0EF5D4C6F800}, to previously placed \textit{JointOrigins}\footnote{https://help.autodesk.com/view/fusion360/ENU/?guid=ASM-JOINT-ORIGIN}.
JointOrigins define a three-dimensional coordinate system with origin in Fusion 360, known as a frame in robotics~\cite[Sec.~1.2]{HandbookOfRobo}, and can be placed by referencing arbitrary geometry.
Many different methods and references can be used to construct JointOrigins. 
As such they are very flexible, arbitrary orientation of the coordinate system axes and arbitrary origin points can be realized.

Annotating types to these JointOrigins links a type to a transformation, allowing applications of combinators to each other to resolve to a transformation in turn.
Each JointOrigin can be annotated with a provided and required type, the presence of one of the two is mandatory. 
This annotation consists of an arbitrary amount of types contained in the three subtype hierarchies.
As previously discussed, this allows expressing what can connect to a JointOrigin.
When first typing a JointOrigin, it is assigned a persistent UUID for technical purposes.
Each JointOrigin with a provided type creates a ``configuration'' of the part.
Configurations represent that a part can be used in different ways.
For instance, consider a wooden cube with different mating geometries on each side. 
Such a cube can \textit{provide} each of the types on its six sides, as long as the \textit{required} types of the other sides are present.
As such it has up to six configurations. 

Additionally, the part itself is also annotated with types from the ``parts'' and ``attributes'' hierarchies. 
These types represent information inherent to the part, for example our previous cube is always a cube, and always made of wood. 
All configurations also provide these types.

\begin{figure}[htbp]
\centerline{\includegraphics[width=0.5\textwidth]{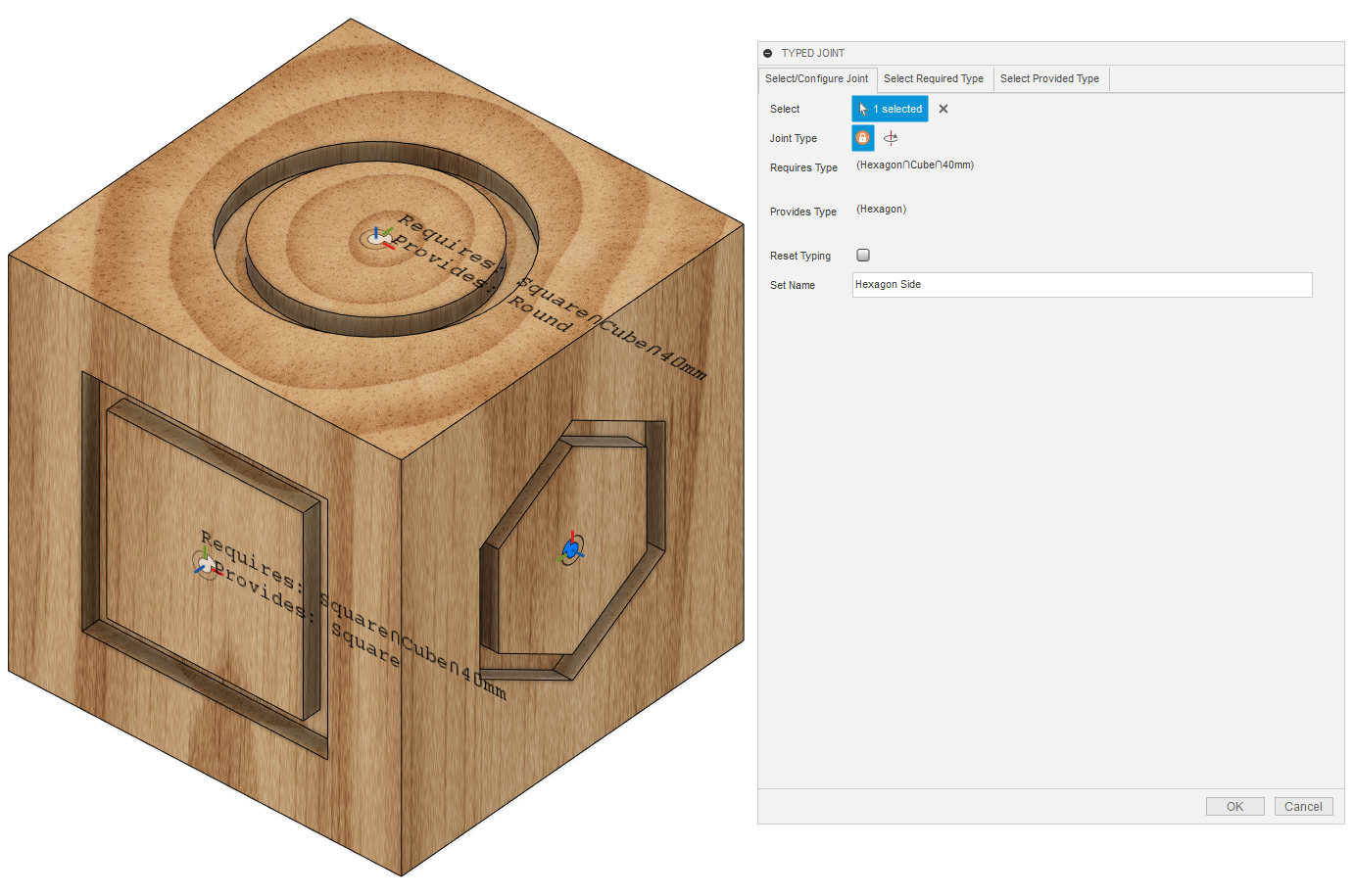}}
\caption{Example of annotating previously mentioned wooden cube with types. Image captured in Fusion 360.}
\label{fig:annotating}
\end{figure}

As seen in \autoref{fig:annotating}, the graphical interface to annotate types to JointOrigins is tightly integrated into Fusion 360. 
The two tabs that are not shown contain the selection interface seen in \autoref{fig:types}.
Selecting multiple JointOrigins enforces that the attached sub-assemblies at these JointOrigins will be identical.

The user is assisted by a graphical user interface that gives feedback on whether the part is sufficiently typed.
The JSON file contains all the type information about the part, alongside some technical attributes to find and manipulate the part in Fusion 360.

\section{Requesting}
Once all required parts have been annotated with types, the user can request sets of assemblies.
To do this there is a graphical user interface that allows selecting a combination of types from the ``parts'' and ``attributes'' subtype hierarchies.
These are used as the request.
Additionally, the user can specify a set of \textit{propagated types}, stemming from all three subtype hierarchies. 
These are used to specify types that are not annotated to the part that is being requested, but could be found elsewhere in the synthesized assembly. 
For instance, when requesting a robotic arm, the user requests a robotic arm base, and by completing the requirements of the base a full arm is obtained. 
However, the base is not aware of what type of effector is present in the later assembly, so we can not specify a robotic arm with a specific effector. 
The propagated types remedy this and allow requesting such specific robotic arms.

Combinators and their corresponding types are computed dynamically based on the request, modifying the combinators' types to correctly propagate according to the user's selection and account for a part's multiple configurations. 
The synthesis request is posed against these typed combinators.
All results offer a limited guarantee of correctness~\cite{Bessai:14} (depending on if the types were correctly annotated and tightly correspond to actual mating geometry).
The first 100 terms of the results are enumerated, post-processed (to refactor the resulting assembly into links, bounded by all non-rigid joints), and persisted. 
It is also possible to request assemblies with specific part counts, in which case the 100 results are divided up across all requested term sizes.
The limit of 100 is an arbitrary default and can be overridden in the request.

\section{Assembling} 
The post-processed results represent simple assembly programs, persisted as JSON files.
They contain a sequence of instructions to be executed by the assembly machine present in our plugin.

\begin{figure}[htbp]
\centerline{\includegraphics[width=0.5\textwidth]{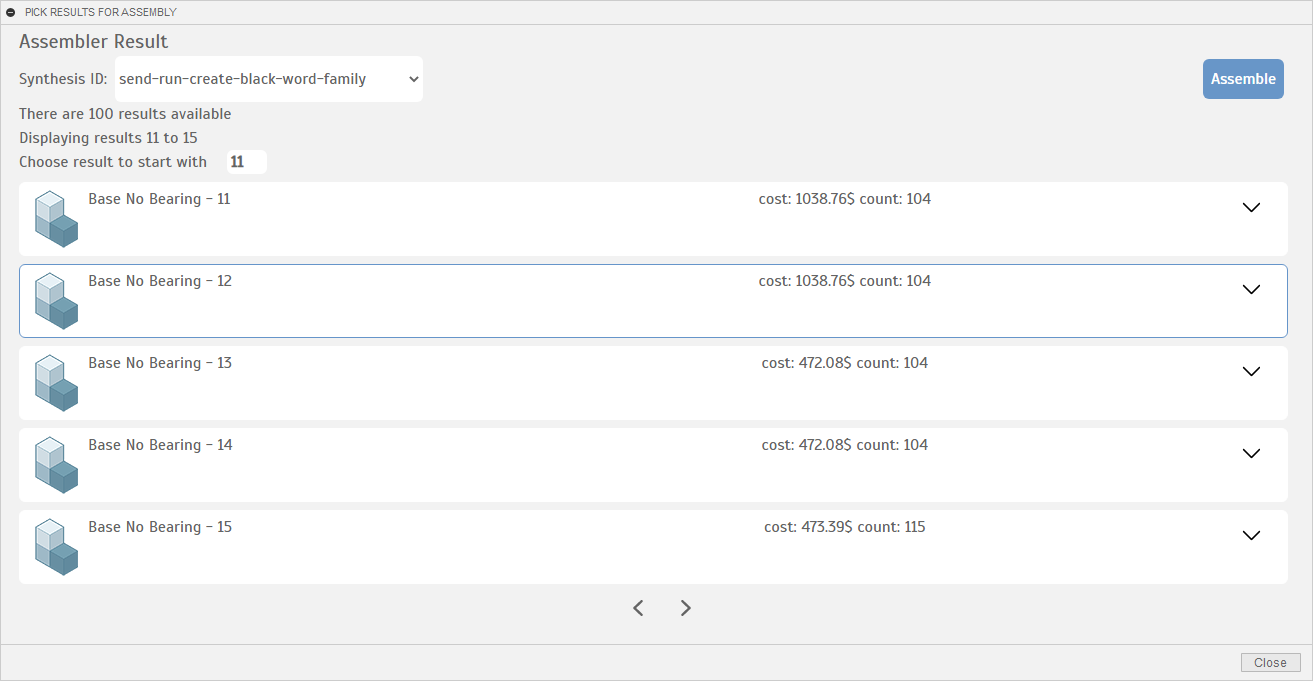}}
\caption{Example of the user interface to browse the synthesized results. }
\label{fig:browse}
\end{figure}

Users can browse results by means of a graphical user interface, pictured in \autoref{fig:browse}. 
Assemblies display their total amount of required parts and their estimated cost (based off of previously set metadata that our plugin can also manage).
Unfolding individual entries shows a full BOM.

When a user selects a result to assemble, as a first step, the instructions tell the plugin which parts to insert, in which order, in which quantities. 
This is important because it greatly speeds up the insertion process in Fusion 360.
This is due to the initial insertion of a part into an assembly being slow, but creating multiple occurrences of it can be optimized to be orders of magnitude faster. 
As a second step, the instructions contain information about how to restructure the parts to move them into separate links. 
Components for the individual links get created, and the inserted parts get moved to the created components. 
The order of these moves is critical, to ensure that the Joints created in the next step correspond to the presence of the parts in the links.

Following this, the actual assembly is built. 
Based on the previously assigned UUIDs, the instructions detail what kind of Joint, rigid or revolute, to create between which JointOrigins.
After all Joints have been created, a complete CAD model of the synthesized assembly is present in Fusion 360. 
It contains functioning links with inverse kinematics.
Apart from manually selecting a model to assemble, all results can also be sequentially assembled in an automated fashion.

\section{Results}
We have modeled a set of 27 parts consisting of different mounting plates, brackets, servo-motors and effectors to illustrate the plugin's functionality. 
These parts are suitable to construct robotic arms from. 
We have created three subtype hierarchies for this set of parts and then annotated the parts with types.
We then posed different requests against the set of parts. 
For these, we verified that requests yields only well-formed results and that requests with propagated types contain parts which provide them.

\begin{figure}[htbp]
\centerline{\includegraphics[width=0.5\textwidth]{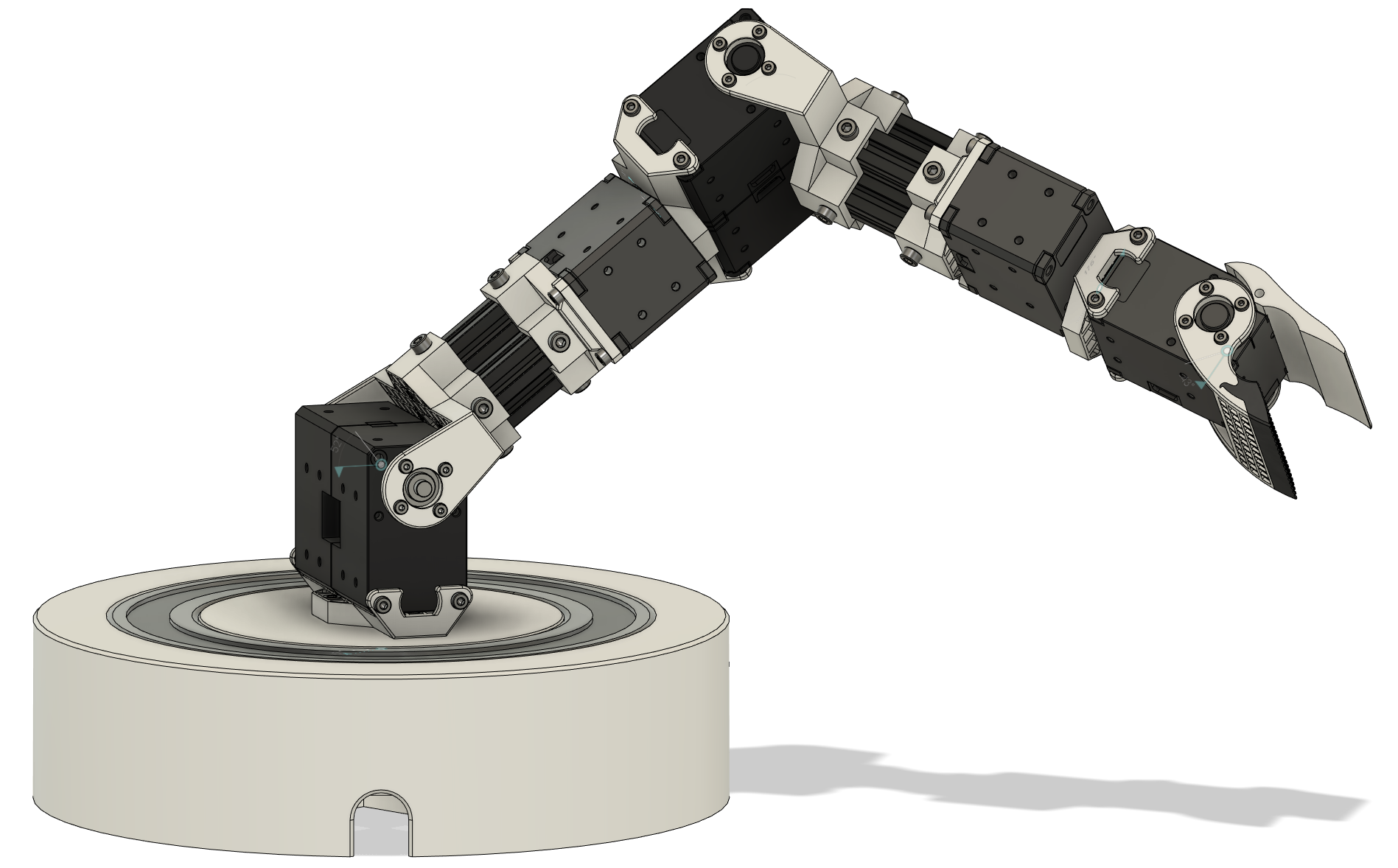}}
\caption{Example of a synthesized 5-DoF robotic arm.}
\label{fig:roboarm}
\end{figure}

To illustrate the effect of propagated types, we propagated information about whether an axis of rotation that revolves a link around itself is present.
This lead to the generated arms exhibiting a higher degree of manipulability. 
An example is shown in \autoref{fig:roboarm}.
The total amount of parts in the arms scaled linearly with their number of links. 
Prices varied greatly between assemblies, mainly due to different motors being used.
This highlights that generating design alternatives can be used to compare different price points and structures. 
In combination with Fusion 360's set of tools, this can be used to identify good designs, without manually creating them. 

\section{Limitations} 
One major issue is that it is difficult to reach a sufficient level of specification. 
Whilst the generated results can be assembled, they contain many results that make little sense in practice due to performance characteristics. 
Additionally, it is difficult to specify numeric performance indicators, i.e.\ a robotic arm with specific degrees of freedom can not be requested.
This is due to a lack of specific domain knowledge in the generated combinators and types. 
This can be remedied by implementing filters on the results, however this is not as desirable as a stronger specification mechanism, which is the focus of ongoing work. 

\section{Conclusion}
CLS-CAD takes a step towards enabling engineers and product designers to enhance the re-usablility of their parts and assemblies. Synthesizing assemblies leaves more time for creative work and improves reliability due to human errors being prevented. Additionally, the design space is more thoroughly explored, as the synthesis is not opinionated. CLS-CAD is still in an early stage, and further augmentations regarding specification, integration and optimization are planned. However, even in this stage it is a powerful tool for applying knowledge-based engineering and combinatory logic to the CAD design process. 

\appendix
A video demonstration/explanation is available at\ \url{https://www.youtube.com/watch?v=gK00StSAxuk}.
The source code is available at GitHub at \url{https://github.com/tudo-seal/CLS-CAD} and \url{https://github.com/tudo-seal/CLS-CPS}.
The data-set and version of the tool used is available online ~\cite{chaumet_constantin_2023_7970609}.

\bibliographystyle{IEEEtran}
\bibliography{IEEEabrv,bibliography.bib}

\end{document}